\documentclass[journal]{IEEEtran}
\usepackage{amsmath,amssymb,amsfonts}
\usepackage{algorithmic}
\usepackage{algorithm}
\usepackage{graphicx}
\usepackage{textcomp}
\usepackage{cite}
\usepackage{bm}
\usepackage{multirow}
\usepackage{booktabs}
\usepackage{color}
\usepackage{acronym}
\acrodef{SIMO}{single-input multiple-output}
\acrodef{CF-mMIMO}{user-centric cell-free massive MIMO}
\acrodef{MIMO}{multiple-input multiple-output}
\acrodef{mMIMO}{massive MIMO}
\acrodef{TDD}{time-division duplex}
\acrodef{CSI}{channel state information}
\acrodef{UE}{user equipment}
\acrodef{BS}{base station}
\acrodef{LOS}{line-of-sight}
\acrodef{GELU}{Gaussian-error linear unit}
\acrodef{NLOS}{non-line-of-sight} 
\acrodef{SIC}{successive interference cancellation}
\acrodef{SNR}{signal-to-noise ratio}
\acrodef{SINR}{signal-to-interference-plus-noise ratio}
\acrodef{GFwd}{Greedy Forward Selection} 
\acrodef{NN}{neural network}
\acrodef{DL}{deep learning}
\acrodef{CDF}{cumulative distribution function}
\acrodef{RIS}{reconfigurable intelligent surface}
\acrodef{RF}{radio frequency}
\acrodef{AWGN}{additive white Gaussian noise}
\acrodef{ULA}{uniform linear array}
\acrodef{FA}{fluid antenna}
\acrodef{MU}{multi-user}
\acrodef{FAS}{fluid antenna system}
\acrodef{SE}{spectral efficiency}
\acrodef{FAMA}{fluid antenna multiple access}
\acrodef{RL}{Reinforcement learning}
\acrodef{IL}{imitation learning}
\acrodef{DC}{digital combining}
\acrodef{GEPort}{generalized eigenvector port selection}
\acrodef{GEV}{generalized eigenvector}
\acrodef{MRC}{maximum ratio combining}
\acrodef{CUMA}{compact ultra massive antenna array}
\acrodef{GFwd+S}{greedy forward selection with swap}

\newcommand{\bH}{\mathbf{H}}
\newcommand{\bA}{\mathbf{A}}
\newcommand{\bB}{\mathbf{B}}
\newcommand{\bC}{\mathbf{C}}
\newcommand{\bS}{\mathbf{S}}
\newcommand{\bZ}{\mathbf{Z}}
\newcommand{\bF}{\mathbf{F}}
\newcommand{\bI}{\mathbf{I}}
\newcommand{\bSigma}{\boldsymbol{\Sigma}}
\newcommand{\bw}{\mathbf{w}}

\newcommand{\bp}{\mathbf{p}}
\newcommand{\bn}{\mathbf{n}}
\newcommand{\bx}{\mathbf{x}}
\newcommand{\be}{\mathbf{e}}
\newcommand{\bh}{\mathbf{h}}
\newcommand{\bff}{\mathbf{f}}
\newcommand{\bzero}{\mathbf{0}}
\newcommand{\cB}{\mathcal{B}}
\newcommand{\cC}{\mathcal{C}}
\newcommand{\cS}{\mathcal{S}}

\newcommand{\cL}{\mathcal{L}}
\newcommand{\cT}{\mathcal{T}}
\newcommand{\cO}{\mathcal{O}}

\begin{document}


\title{Greedy and {Transformer}-Based {Multi-Port} Selection for {Slow} Fluid Antenna Multiple Access}

\author{Darian P\'erez-Ad\'an,~\IEEEmembership{Member,~IEEE,} Jos\'e P. Gonz\'alez-Coma, F. Javier L\'opez-Mart\'inez,~\IEEEmembership{Senior Member,~IEEE} and Luis Castedo,~\IEEEmembership{Senior Member,~IEEE}%
\thanks{This work has been supported by grant ED431C 2024/18 funded by Xunta de Galicia, by grant PICUD-2025-02 (COMTEUM) funded by the Defense University Center at the Spanish Naval Academy, by grants PID2022-137099NB-C42 (MADDIE) and PID2023-149975OB-I00 (COSTUME) funded by MICIU/AEI/10.13039/501100011033 and FEDER/UE, and by the postdoctoral Grant No. ED481B-2025/092 funded by Xunta de Galicia.}%
\thanks{D. P\'{e}rez-Ad\'{a}n and L. Castedo are with the Department of Computer Engineering, University of A Coru\~{n}a, CITIC, A Coru\~{n}a, Spain, e-mail: \{d.adan, luis\}@udc.es.}%
\thanks{J.P. Gonz\'{a}lez-Coma is with the Defense University Center at the Spanish Naval Academy, Mar\'{\i}n, Spain, email: jose.gcoma@cud.uvigo.es.}%
\thanks{F.J. L\'{o}pez-Mart\'{\i}nez is with the Dept. Signal Theory, Networking and Communications, Research Centre for Information and Communication Technologies (CITIC-UGR), University of Granada, 18071, Granada, Spain. e-mail: fjlm@ugr.es.}%
}

\maketitle
\markboth{IEEE WIRELESS COMMUNICATIONS LETTERS}%
{P\'erez-Ad\'an \MakeLowercase{\textit{et al.}}: Greedy and Learning-Based Port Selection for Fluid Antenna Multiple Access with Multi-Port Receivers}
\begin{abstract}
We address the port-selection problem in \ac{FAMA} systems with multi-port \ac{FA} receivers. Existing methods either achieve near-optimal \ac{SE} at prohibitive computational cost or sacrifice significant performance for lower complexity. We propose two complementary strategies: (i) GFwd+S, a greedy forward-selection method with swap refinement that consistently outperforms state-of-the-art reference schemes in terms of \ac{SE}, and (ii) a Transformer-based neural network trained via imitation learning followed by a Reinforce policy-gradient stage, which approaches GFwd+S performance at lower computational cost. 


\end{abstract}

\begin{IEEEkeywords}
Fluid antenna systems, multi-port selection, Reinforcement learning, greedy algorithms, MIMO, multi-user.
\end{IEEEkeywords}

\section{Introduction}\label{sec:intro}

\Acp{FAS} are emerging as a promising alternative to conventional \ac{MIMO} systems, which rely on fixed-position antenna arrays \cite{11247926}. By dynamically selecting one among many densely packed port positions within a compact aperture, \acp{FAS} leverages fine-grained spatial diversity to enhance beamforming gains and improve signal reception \cite{9131873}. A key application is fluid antenna multiple access (FAMA) \cite{9650760}, which enables open-loop multiple access with \ac{CSI} required only at the receiver. The slow-FAMA paradigm \cite{wong2023slowfama} relaxes the stringent port-switching requirements of fast-FAMA, reducing complexity while still allowing user multiplexing.

The slow-FAMA framework has been extended to enable multi-port selection using $L>1$ \ac{RF} chains \cite{wong2024cuma,gonzalezcoma2026slowfama,Hong2025}.
Although exhaustive search over all port subsets is optimal, it is computationally prohibitive in practice. Hence, heuristic schemes such as \ac{CUMA} were first proposed \cite{wong2024cuma}. More recently, \cite{gonzalezcoma2026slowfama} proposed a joint design of the port-selection matrix and digital combining vector via iterative backward elimination based on the \ac{GEV} structure of the signal and interference matrices. This work provided the first theoretically grounded approach to multi-port selection in FAMA, achieving a remarkable performance gain even for small $L$, at the expense of cubic complexity in the number of ports. 

Lower-complexity alternatives such as \ac{DC}~\cite{gonzalezcoma2026slowfama} 
and the greedy incremental strategy in~\cite{Hong2025} reduce the computational burden, but still suffer from important limitations. Similar to \ac{CUMA}, \ac{DC} incurs a significant \ac{SE} loss, while the forward-only construction in~\cite{Hong2025} is sensitive to the initial selections and cannot recover from suboptimal early choices. In addition, none of these methods leverages learning to exploit the statistical structure across channel realizations, despite the demonstrated potential of learning-based approaches for antenna selection in conventional \ac{MIMO}~\cite{joung2021ras_dnn}. In the \ac{FAMA} context \cite{waqar2023dl_fama} proposed a deep \ac{NN}-based scheme for single-port selection from partial observations, but its extension to multi-port receivers with combinatorial selection and \ac{GEV} combining remains unexplored.

In this letter, we make two main contributions. First, we
propose \ac{GFwd}, a forward greedy algorithm that incrementally selects ports by maximizing the \ac{SINR} gain, achieving higher \ac{SE} than \ac{GEPort} \cite{gonzalezcoma2026slowfama} at lower complexity. A swap-based refinement step, termed GFwd+S, is further introduced to avoid local optima and improve performance. Second, to reduce complexity further, we design a Transformer-based \ac{NN} trained via \ac{IL} followed by a Reinforce policy-gradient stage, which approaches near-optimal \ac{SE} performance with significantly lower inference latency than both \ac{GEPort} and GFwd+S.

\textit{Notation:} Boldface lowercase ($\mathbf{a}$) and uppercase ($\mathbf{A}$) letters denote vectors and matrices, respectively. Transpose and conjugate transpose are denoted as $(\cdot)^T$ and $(\cdot)^H$. Calligraphic letters, e.g., $\cS$, denote sets, and $|\cS|$ is the set cardinality. $\bI_P$ is the $P \times P$ identity matrix. Finally, $\mathbb{E}\{\cdot\}$ is the expectation operator and $\|\cdot\|_p$ is the $\ell_p$-norm.

\section{System Model}\label{sec:system}

We consider a \ac{BS} with $N_{\text{t}}$ antennas serving $K$ single-antenna users, where each user is equipped with a \ac{FA} array with $P$ ports and $L>1$ \ac{RF} chains to activate multiple \ac{FA} ports. Following the slow-FAMA paradigm~\cite{wong2023slowfama}, we set $N_{\text{t}} = K$, and the \ac{BS} uses canonical precoding vectors $\bp_k = \be_k$, as in \cite{gonzalezcoma2026slowfama}, requiring no \ac{CSI} at the transmitter\footnote{The same \ac{CSI} availability is assumed at all receivers. Channel estimation for \acp{FA} has been studied in~\cite{New2025oversampling, New2025training}.}.

The received signal at the $k$-th user is
\begin{equation}\label{eq:rxsignal}
\bx_k = \bH_k \bp_k z_k + \sum_{j \neq k} \bH_k \bp_j z_j + \bn_k,
\end{equation}
where $\bH_k \in \mathbb{C}^{P \times N_{\text{t}}}$ is the channel matrix between the \ac{BS} and the $k$-th user, $z_k \in \mathbb{C}$ is the data symbol with $\mathbb{E}\{|z_k|^2\} = \sigma_{\text{S}}^2$, and $\bn_k \in \mathbb{C}^{P \times 1}$ is the \ac{AWGN} vector with per-element power~$\sigma_{\text{n}}^2$. At the receiver, a port selection matrix $\bS_k \in \cB$, where $\cB := \{\bZ \in \{0,1\}^{P \times L} : \|\bZ\|_{0,\infty} \leq 1\}$ selects $L$ active ports, while a combining vector $\bw_k \in \mathbb{C}^{L}$ satisfying $\|\bw_k\|_2 = 1$ yields the estimated symbol
\begin{equation}\label{eq:rxsymbol}
\hat{z}_k = \bw_k^{H} \bS_k^{T} \bx_k.
\end{equation}
We adopt Jakes' correlation model for a 1D \ac{FA}~\cite{ramirez2024spatial}, under which the columns of $\bH_k$ are i.i.d. and distributed as $\mathcal{CN}(\bzero, \bSigma_k)$, where
\begin{equation}\label{eq:corr}
[\bSigma_k]_{p,p'} = \mathrm{sinc}\big(2(d_p - d_{p'})\big),
\end{equation}
and $d_p = (p-1)W/(P-1)$ denotes the normalized position of the $p$-th port within a \ac{FA} of size~$W\lambda$. The performance metric considered in this work is the \ac{SE}, given for user $k$ by $R_k = \log_2(1 + \mathrm{SINR}_k)$, where the \ac{SINR} is defined as
\begin{equation}\label{eq:sinr}
\mathrm{SINR}_k = \frac{\big|\bw_k^{H} \bS_k^{T} \bH_k \bp_k\big|^2}{\sum_{j \neq k}\big|\bw_k^{H} \bS_k^{T} \bH_k \bp_j\big|^2 + \frac{1}{\mathrm{SNR}}},
\end{equation}
with $\mathrm{SNR} = \sigma_{\text{S}}^2 / \sigma_{\text{n}}^2$ denoting the transmit \ac{SNR}. Accordingly, the optimization problem is formulated as
\begin{equation}\label{eq:problem}
\max_{\{\bS_k \in \cB,\, \bw_k \in \mathbb{C}^{L},\|\bw_k\|_2 = 1\}_{k=1}^K} \sum\nolimits_{k=1}^{K} \log_2\!\left(1 + \mathrm{SINR}_k\right).
\end{equation}
For a given $\bS_k$, the optimal combiner $\bw_k$ is the dominant \ac{GEV} of the matrix pair $(\tilde{\bA}_k, \tilde{\bB}_k)$~\cite{gonzalezcoma2026slowfama,schubert2004multiuser}, where
\begin{equation}\label{eq:AkBk}
\tilde{\bA}_k = \bS_k^{T} \bA_k \bS_k, \quad \tilde{\bB}_k = \bS_k^{T} \bB_k \bS_k,
\end{equation}
with the signal matrix defined as $\bA_k = \bH_k \bp_k \bp_k^{H} \bH_k^{H}$ and the interference-plus-noise matrix as $\bB_k = \sum_{j \neq k} \bH_k \bp_j \bp_j^{H} \bH_k^{H} + \bI_P / \mathrm{SNR}$.

\section{Proposed Port Selection Methods}\label{sec:proposed}
The design of the port selection matrix $\bS_k$ is challenging because it affects both the desired signal and the interference. Moreover, an exhaustive search over all $\binom{P}{L}$ possible subsets is computationally prohibitive for practical values of $P$ and $L$. In this context, the \ac{GEPort} algorithm~\cite{gonzalezcoma2026slowfama} provides strong performance through backward elimination, but it requires $P - L$ eigen-decompositions on progressively smaller matrices, resulting in $\cO(P^3)$ complexity\footnote{This is a conservative upper bound obtained by assigning a cost of $P^3$ to each of the $P-L$ decompositions; the exact complexity is $\cO(P^4)$ for $L \ll P$~\cite{gonzalezcoma2026slowfama}.}. To reduce this complexity, we propose two complementary strategies offering different performance--complexity trade-offs.

\subsection{\ac{GFwd} with Swap Refinement}\label{sec:gfwd}

In contrast to \ac{GEPort}~\cite{gonzalezcoma2026slowfama}, which starts from all $P$ ports and iteratively removes the least contributing one, we build the selection set incrementally\footnote{An incremental strategy with a \textit{fixed} covariance-based interference rejection vector was considered in \cite{Hong2025}.}. Starting from $\cS = \emptyset$ and $\cC = \{1, \ldots, P\}$, at each step $t = 1, \ldots, L$, we add the port that maximizes the \ac{SINR}:
\begin{equation}\label{eq:gfwd_step}
p^{*} = \arg\max_{p \in \cC} \; \lambda_{\max}\!\Big(\tilde{\bA}_{\cS \cup \{p\}}, \; \tilde{\bB}_{\cS \cup \{p\}}\Big),
\end{equation}
where $\lambda_{\max}(\cdot, \cdot)$ denotes the dominant \ac{GEV}. 
Since \ac{GFwd} operates on matrices of increasing size $t \times t$, for $t=1, \ldots L$, and evaluates up to $P - t + 1$ candidates at step $t$, its total complexity is
\begin{multline}\label{eq:gfwd_complexity}
\sum_{t=1}^{L}(P-t+1)\,t^3 \approx P\sum_{t=1}^{L} t^3 
\overset{(a)}{=} P \cdot \frac{L^2(L+1)^2}{4} \\ =\cO(PL^4), \quad L \ll P,
\end{multline}
where $(a)$ follows from the sum-of-cubes formula. Therefore, the complexity $\cO(PL^4)$ is substantially lower than that of \ac{GEPort} for $L \ll P$. Interestingly, \ac{GFwd} is guaranteed to produce non-decreasing \ac{SINR} values at each incremental step, as shown in Appendix~\ref{app:monotonicity}.

After \ac{GFwd} converges, we perform a local swap refinement to escape local optima. For each selected port $p_i \in \cS$ and each candidate port $p' \in \cC$, we evaluate the \ac{SINR} of $(\cS \setminus \{p_i\}) \cup \{p'\}$ and apply the best improving swap. This procedure is repeated for at most $R$ rounds, or until no further improvement is found, with an additional complexity of $\cO(R P L^4)$. The complete GFwd+S procedure is summarized in Algorithm~\ref{alg:gfwd}.

\begin{algorithm}[htbp]
\caption{Greedy Forward Selection with Swap (GFwd+S)}
\label{alg:gfwd}
\begin{algorithmic}[1]
\REQUIRE $\bA, \bB \in \mathbb{C}^{P \times P}$, number of ports $L$, max rounds $R$
\STATE $\cS \leftarrow \emptyset$, \; $\cC \leftarrow \{1, \ldots, P\}$
\FOR{$t = 1$ \TO $L$}
  \STATE $p^{*} \leftarrow \arg\max_{p \in \cC} \lambda_{\max}\!\big(\tilde{\bA}_{\cS \cup \{p\}}, \tilde{\bB}_{\cS \cup \{p\}}\big)$
  \STATE $\cS \leftarrow \cS \cup \{p^{*}\}$, \; $\cC \leftarrow \cC \setminus \{p^{*}\}$
\ENDFOR
\STATE $\gamma^{*} \leftarrow \lambda_{\max}\!\big(\tilde{\bA}_{\cS}, \tilde{\bB}_{\cS}\big)$
\FOR{$r = 1$ \TO $R$}
  \STATE $\text{improved} \leftarrow \text{false}$
  \FOR{each $p_i \in \cS$}
    \STATE $\cT \leftarrow \cS \setminus \{p_i\}$
    \STATE $\hat{p} \leftarrow \displaystyle\arg\max_{p' \in \cC}\; \lambda_{\max}\!\big(\tilde{\bA}_{\cT \cup \{p'\}},\, \tilde{\bB}_{\cT \cup \{p'\}}\big)$
    \IF{$\lambda_{\max}\!\big(\tilde{\bA}_{\cT \cup \{\hat{p}\}},\, \tilde{\bB}_{\cT \cup \{\hat{p}\}}\big) > \gamma^{*}$}
      \STATE $\cS \leftarrow \cT \cup \{\hat{p}\}$, \; $\cC \leftarrow (\cC \setminus \{\hat{p}\}) \cup \{p_i\}$
      \STATE Update $\gamma^{*}$, \; $\text{improved} \leftarrow \text{true}$
    \ENDIF
  \ENDFOR
  \IF{not improved} \STATE \textbf{break} \ENDIF
\ENDFOR
\RETURN $\cS$, \; $\bw = $ dominant eigenvector of $(\tilde{\bA}_{\cS}, \tilde{\bB}_{\cS})$
\end{algorithmic}
\end{algorithm}

\subsection{Transformer-Based Neural Port Selection}\label{sec:nn}
\begin{figure}[t]
  \centering
  \vspace*{-3mm}
  \includegraphics[width=\columnwidth]{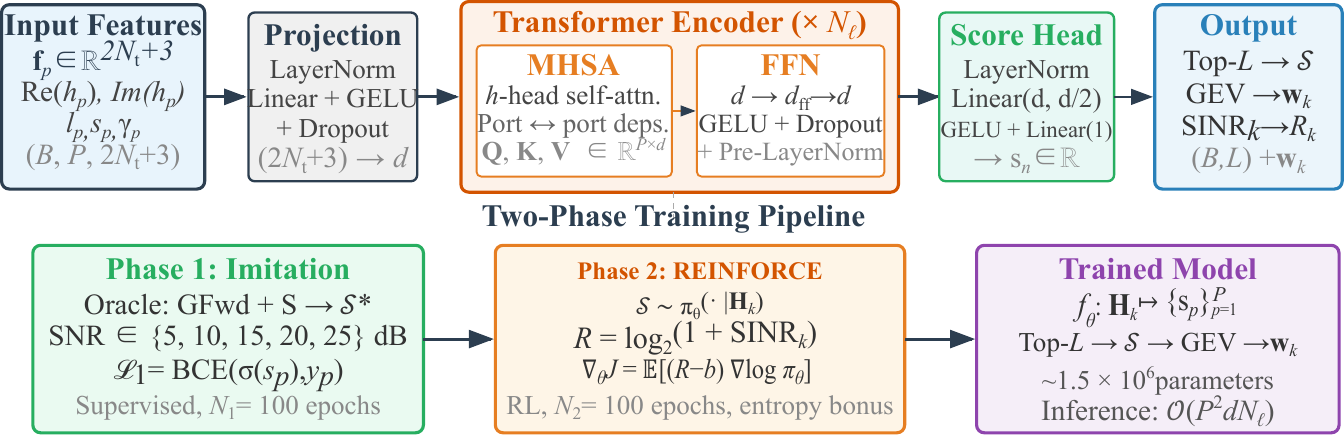}
    \vspace*{-5mm}
  \caption{Transformer-based NN port selector architecture: 2 training phases.}
  \label{fig:arch}
\end{figure}

As shown later, GFwd+S achieves higher \ac{SE} than competing schemes. However, its computational cost motivates a learning-based alternative for low-complexity port selection across channel realizations. We therefore propose a data-driven method based on a Transformer encoder~\cite{vaswani2017attention} that scores all $P$ ports simultaneously and captures inter-port dependencies through self-attention (see Fig.~\ref{fig:arch}). Let $f_\theta \colon \mathbb{C}^{P \times N_{\text{t}}} \to \mathbb{R}^P$ denote the \ac{NN} mapping from the channel $\bH_k$ to the score vector 
$\mathbf{s} = f_\theta(\bH_k)$, where $\theta$ denotes the trainable parameters.

\subsubsection{Input Features}
For user $k$ (without loss of generality), per-port features are extracted from $\bH_k \in \mathbb{C}^{P \times N_{\text{t}}}$ as
\begin{equation}\label{eq:features}
\bff_p = \big[\mathrm{Re}(\tilde{\bh}_p^{T}),\; 
\mathrm{Im}(\tilde{\bh}_p^{T}),\; \bar{\gamma}_p,\; 
\bar{s}_p,\; \bar{\iota}_p \big] \in \mathbb{R}^{2N_{\text{t}}+3},
\end{equation}
where $\tilde{\bh}_p = \bh_p / \|\bH_k\|_F$ denotes the $p$-th row of the Frobenius-normalized channel matrix, and $\bar{\gamma}_p$, $\bar{s}_p$, $\bar{\iota}_p$ denote the per-port \ac{SINR}, signal, and interference values, respectively, normalized by their corresponding port-wise maxima.

\subsubsection{Architecture}
The feature matrix $\bF \in \mathbb{R}^{P \times (2N_{\text{t}}+3)}$ is first projected onto a $d$-dimensional space through LayerNorm followed by a linear layer with \ac{GELU} activation. The resulting sequence is then processed by a stack of $N_\ell$ Transformer encoder layers with $h$-head self-attention. Self-attention captures pairwise inter-port dependencies in $\cO(P^2)$ operations, regardless of spatial separation, which is particularly important under the spatially correlated \ac{FA} channel model. A scoring head maps each token to a scalar $s_p \in \mathbb{R}$, and the top-$L$ ports are selected. The \ac{GEV} combiner then computes the optimal $\bw$ for the selected subset. Dropout is applied after each sublayer for regularization. The inference complexity is $\cO(P^2 d N_\ell)$, dominated by the self-attention operation.

\subsubsection{Two-Phase Training}
Direct \ac{RL} training from scratch is unstable due to the large combinatorial action space. In contrast, pure supervised learning via \ac{IL} converges quickly but is limited by the cross-entropy loss, which does not directly optimize \ac{SE}. We therefore combine both: \ac{IL} provides a warm-start initialization close to the GFwd+S oracle, and Reinforce subsequently fine-tunes the policy to maximize \ac{SE} directly.

\textbf{Phase~1 (\ac{IL}):} A labeled dataset $\{(\bH^{(i)}, \cS^{*(i)})\}$ of $15{,}000$ training and $1{,}000$ validation samples is generated using GFwd+S as the oracle over \ac{SNR} values $\{5, 10, 15, 20, 25\}$~dB. The \ac{NN} is trained to predict the oracle-selected ports via the binary cross-entropy loss
\begin{equation}\label{eq:bce}
\cL_1 = -\frac{1}{P}\sum_{p=1}^{P} \big[y_p \log \sigma(s_p) 
+ (1 - y_p)\log\big(1 - \sigma(s_p)\big)\big],
\end{equation}
where $y_p = 1$ if $p \in \cS^{*}$ and $\sigma(\cdot)$ is the sigmoid function.

\textbf{Phase~2 (Reinforce):} Starting from the \ac{IL}-trained parameters ($\theta$), we use the Reinforce policy gradient~\cite{williams1992Reinforce} to directly maximize the \ac{SE}. The \ac{NN} sequentially samples $L$ ports from $\pi_\theta(\cS \mid \bH)$ without replacement, renormalizing the categorical distribution after each draw. The policy gradient is
\begin{equation}\label{eq:Reinforce}
\nabla_\theta J(\theta) = \mathbb{E}\Big[\big(R - b\big)\, 
\nabla_\theta \log \pi_\theta(\cS \mid \bH)\Big],
\end{equation}
where $R = R_k(\cS)$ is the \ac{SE} with \ac{GEV} combining, and $b$ is an exponential moving-average baseline. An entropy bonus is added to promote exploration during training. The overall pipeline is summarized in Algorithm~\ref{alg:nn_training}. The inference complexities of \ac{GFwd}, GFwd+S, and the \ac{NN} are $\cO(PL^4)$, $\cO(RPL^4)$, and $\cO(P^2 d N_\ell)$, respectively; measured execution times are reported in Section~\ref{sec:results}.

\begin{algorithm}[t]
\caption{\ac{NN} Training Pipeline}
\label{alg:nn_training}
\begin{algorithmic}[1]
\REQUIRE GFwd+S oracle, \ac{SNR} range, epochs $N_1$, $N_2$
\STATE \textbf{Phase 1:} Generate $\{(\bH^{(i)}, \cS^{*(i)})\}$ via GFwd+S
\FOR{epoch $= 1$ \TO $N_1$}
  \STATE Update $\theta$ by minimizing $\cL_1$ in~\eqref{eq:bce}
\ENDFOR
\STATE \textbf{Phase 2:}
\FOR{epoch $= 1$ \TO $N_2$}
  \STATE Sample \ac{SNR}, generate $\bH$, sample $\cS \sim \pi_\theta$
  \STATE Compute $R = \log_2(1 + \mathrm{SINR}(\cS))$ via \ac{GEV}
  \STATE Update $\theta$ via~\eqref{eq:Reinforce} with entropy bonus
\ENDFOR
\RETURN Trained parameters $\theta$ 
\end{algorithmic}
\end{algorithm}

\section{Numerical Results}\label{sec:results}
In this section, we evaluate the proposed methods through simulations under different system setups and analyze their computational complexity. Unless otherwise indicated, the simulation parameters are given in Table~\ref{tab:setup}. Phase~1 uses labeled samples generated by GFwd+S over \ac{SNR} values $\{5, 10, 15, 20, 25\}$~dB for $N_1 = 100$ epochs. Phase~2 runs Reinforce for $N_2 = 100$ epochs, with the \ac{SNR} sampled uniformly from $[5, 27]$~dB. For benchmarking, we compare the proposed methods against slow-FAMA~\cite{wong2023slowfama}, which selects the single best port per user; \ac{DC}~\cite{gonzalezcoma2026slowfama}, which extends slow-FAMA by selecting the $L$ ports with the highest individual \ac{SINR} values and then applying \ac{GEV} combining; \ac{CUMA}~\cite{wong2024cuma}, which phase-aligns a subset of ports for constructive combining; and \ac{GEPort}~\cite{gonzalezcoma2026slowfama}, which jointly designs the selection matrix and combiner through iterative backward elimination.
\begin{table}[htbp]
\centering
\caption{Simulation Parameters}
\label{tab:setup}
\renewcommand{\arraystretch}{1.1}
\begin{tabular}{@{}lc|lc@{}}
\toprule
\textbf{Parameter} & \textbf{Value} & \textbf{Parameter} & \textbf{Value} \\
\midrule
$P$            & 100        & $d$~(model dim.) & 192 \\
$L$            & 8         & $h$~(heads)      & 6 \\
$K = N_{\text{t}}$        & 10         & $N_\ell$~(layers) & 5 \\
$W$            & $4\lambda$ & $d_{\mathrm{ff}}$ & 384 \\
Correlation    & Jakes~\cite{ramirez2024spatial} & Dropout & 0.05 \\
\bottomrule
\end{tabular}
\end{table}

\subsection{Computational Complexity Analysis}\label{sec:complexity}

Table~\ref{tab:complexity} summarizes the computational complexity and inference times of all considered methods, including the low-complexity baselines (Slow-FAMA, \ac{DC}, and \ac{CUMA}). For the proposed \ac{NN}, the forward pass has complexity $\cO(P^2 d N_\ell)$, dominated by self-attention, followed by the \ac{GEV} combining with complexity $\cO(L^3)$. In contrast, \ac{GEPort} requires $\cO\big((P{-}L)P^3\big)$ due to repeated eigenvalue problems, while GFwd+S has complexity $\cO(RPL^4)$ due to swap refinement.

\begin{table}[t]
\centering
\caption{Computational Complexity and Inference Times}
\label{tab:complexity}
\renewcommand{\arraystretch}{1.15}
\begin{tabular}{@{}lcc@{}}
\toprule
\textbf{Method} & \textbf{Complexity} & \textbf{Time (ms)} \\
\midrule
Slow-FAMA~\cite{wong2023slowfama}     & $\cO(PK)$              & 0.28 \\
\ac{DC}~\cite{gonzalezcoma2026slowfama}     & $\cO(PK + L^3)$        & 0.42 \\
\ac{CUMA}~\cite{wong2024cuma}               & $\cO(P)$               & 0.15 \\
\ac{GFwd} (prop.)                           & $\cO(PL^4)$            & 90.90 \\
GFwd+S (prop.)                         & $\cO(RPL^4)$          & 385.13 \\
\ac{GEPort}~\cite{gonzalezcoma2026slowfama} & $\cO\big((P{-}L)P^3\big)$ & 232.04 \\
\ac{NN} (prop.)                             & $\cO(P^2 d N_\ell)$    & 1.53 \\
\bottomrule
\multicolumn{3}{l}{\footnotesize Setup: $P{=}100$, $L{=}8$, $K{=}N_{\text{t}}{=}10$, $R=3$. HW: Intel Core Ultra~7}\\
\multicolumn{3}{l}{\footnotesize (16c, 3.8\,GHz), 32\,GB RAM, 8\,GB GPU. SW: Python~3.13, PyTorch~2.6.}\\
\multicolumn{3}{l}{\footnotesize $^*$GFwd+S latency exceeds \ac{GEPort} for $R=3$; \ac{GFwd} alone is faster.}\\
\end{tabular}
\end{table}

\subsection{SE Performance and Scalability Analysis}

\begin{figure}[htbp]
\centering
\vspace{-4mm}
\includegraphics[width=.90\columnwidth]{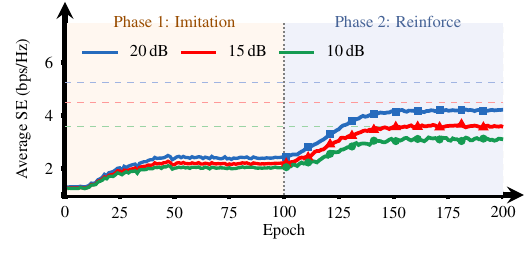}
\vspace{-4mm}
\caption{Validation \ac{SE} during training at \ac{SNR}~$\in \{10, 15, 20\}$~dB.}
\label{fig:train}
\end{figure}

Fig.~\ref{fig:train} shows the training convergence. During \ac{IL} (Phase~1), the validation \ac{SE} at three \ac{SNR} levels increases gradually and saturates at about half of the oracle \ac{SE}. After switching to Reinforce (Phase~2), the \ac{SE} rises steeply---nearly doubling at 20~dB---and stabilizes within roughly 50 epochs. This gain is more pronounced at high \ac{SNR}, where port selection becomes increasingly important relative to noise, thereby making the reward signal more informative for policy optimization.

\begin{figure}[t]
\centering
\vspace{-4mm}
\includegraphics[width=.90\columnwidth]{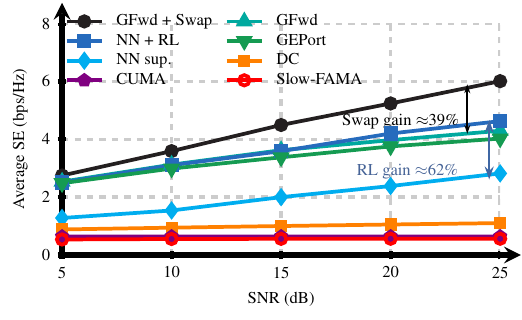}
\vspace{-5mm}
\caption{Average \ac{SE} versus \ac{SNR} for $P=100$, $L=8$, and $K=10$.}
\label{fig:se_vs_snr}
\end{figure}

Fig.~\ref{fig:se_vs_snr} presents the average \ac{SE} versus the transmit \ac{SNR}. Standalone \ac{GFwd} slightly improves upon \ac{GEPort} while requiring lower complexity. With swap refinement, GFwd+S consistently outperforms \ac{GEPort} by up to roughly 40\%, confirming that swap refinement is essential for incremental forward construction to overcome the suboptimal early decisions inherent to backward elimination. Recall that Proposition~1 in Appendix \ref{app:monotonicity} guarantees non-decreasing \ac{SINR} values for \ac{GFwd} at each step. In contrast, \ac{GEPort} only approximates the \ac{SINR} degradation caused by port removal. Since the matrix dimensionality is reduced sequentially and aggressively, early decisions may become suboptimal as the elimination proceeds, which explains the consistent \ac{SE} advantage of the \ac{GFwd}-based methods in Fig.~\ref{fig:se_vs_snr}. The \ac{NN} trained with Reinforce (NN+RL) achieves up to a 62\% gain over the baseline \ac{NN} at high \ac{SNR}, matches or exceeds {GEPort/GFwd} for $\mathrm{SNR} \geq 15$~dB, and reaches over 77\% of GFwd+S (upper bound) performance across all operating points. The low-complexity baselines (slow-FAMA, \ac{DC}, \ac{CUMA}) remain well below, confirming the need for intelligent port selection.

\begin{figure}[htbp]
\centering
\vspace*{-3mm}
\includegraphics[width=.90\columnwidth]{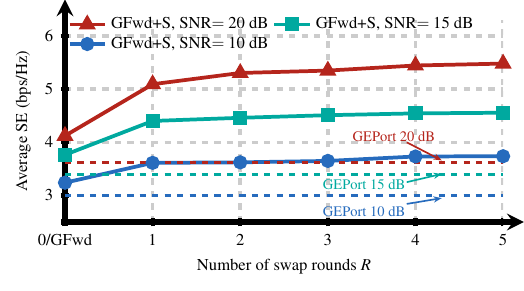}
\vspace*{-5mm}
\caption{Average \ac{SE} versus swap rounds $R$ for 
$P=100$, $L=8$, and $K=10$. Dashed lines indicate \ac{GEPort} 
reference performance at each \ac{SNR}.}
\label{fig:se_vs_swap}
\end{figure}

Fig.~\ref{fig:se_vs_swap} shows the average \ac{SE} versus the number of swap rounds $R$ for \ac{SNR} values in $\{10, 15, 20\}$~dB. A single swap round recovers most of the gain over \ac{GFwd}, with only marginal improvement beyond $R=2$. In addition, GFwd+S consistently outperforms \ac{GEPort} for all considered \ac{SNR} values and $R \geq 1$, confirming that a small number of swap rounds (e.g., $R=3$) is enough to converge at a stable solution.

\begin{figure}[htbp]
\centering
\vspace*{-6mm}
\includegraphics[width=.90\columnwidth]{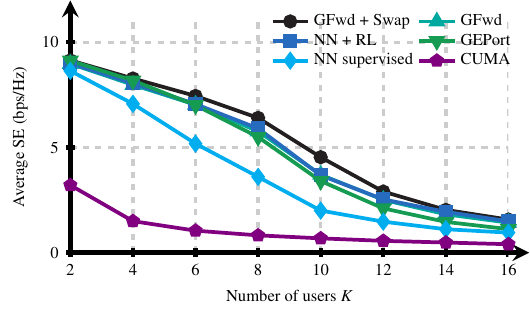}
\vspace*{-5mm}
\caption{Average \ac{SE} vs. $K$ (users) for $P=100$, $L=8$, and $\mathrm{SNR}=15$~dB.}
\label{fig:se_vs_K}
\end{figure}

Fig.~\ref{fig:se_vs_K} shows the \ac{SE} versus the number of users $K$. As $K$ increases, all methods degrade due to the growing inter-user interference. Nevertheless, GFwd+S and the proposed \ac{NN} keep their relative gains, with the \ac{NN} closely approaching GFwd+S. For $K>12$, the performance for all schemes drops sharply because of the strong inter-user interference.

\begin{figure}[htpb]
\centering
\vspace*{-3mm}
\includegraphics[width=.90\columnwidth]{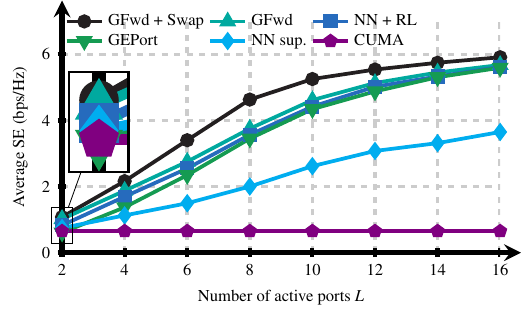}
\vspace*{-4mm}
\caption{Average \ac{SE} vs. RF chains $L$ for $P=100$, $K=10$, $\mathrm{SNR}=15$~dB.}
\label{fig:se_vs_L}
\end{figure}

Fig.~\ref{fig:se_vs_L} depicts the \ac{SE} versus the number of active ports $L$. All methods benefit from increasing $L$ due to additional combining gain. GFwd+S leads to the highest \ac{SE}, while the proposed \ac{NN} outperforms \ac{GEPort} for $L \geq 8$.  The gap between GFwd+S and \ac{GEPort} is largest for intermediate values of $L$ (6--12), where port selection is most combinatorial, and narrows for very small or very large $L$.

\begin{figure}[htpb]
\centering
\vspace{-3mm}
\includegraphics[width=.90\columnwidth]{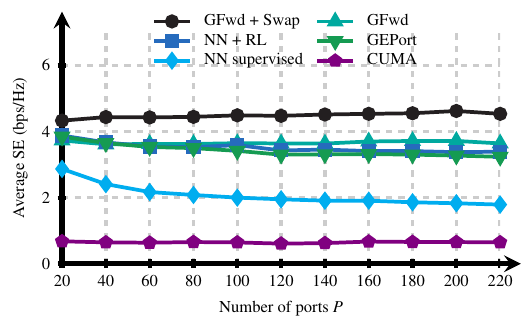}
\vspace{-5mm}
\caption{Average \ac{SE} versus total ports $P$ for $K=10$, $L=8$, $\mathrm{SNR}=15$~dB.}
\label{fig:se_vs_P}
\end{figure}

Fig.~\ref{fig:se_vs_P} shows the \ac{SE} versus the total number of ports $P$ at $\mathrm{SNR}=15$~dB. Increasing $P$ with fixed aperture $W=4\lambda$ densifies the port grid and increases the spatial correlation among ports. In this regime, CUMA, which does not jointly account for $\bA_k$ and $\bB_k$ degrades, while GFwd+S and the proposed \ac{NN} maintain their advantage, as observed in \cite{gonzalezcoma2026slowfama}. The \ac{NN} also consistently outperforms \ac{GEPort} across all $P$ values, achieving more than 75\% of the GFwd+S \ac{SE}.

\vspace{-0.2cm}
\section{Conclusions}\label{sec:conclusion}

We proposed two port selection strategies for multi-port \ac{FAMA} receivers, each addressing complementary aspects of the performance--complexity trade-off. \ac{GFwd} with swap refinement achieves the highest \ac{SE} among all considered methods by avoiding the suboptimal early decisions inherent to backward elimination, as formally supported by the monotonicity property proved in Appendix~\ref{app:monotonicity}. The proposed Transformer-based \ac{NN}, trained via \ac{IL} followed by Reinforce, bridges the gap between low-latency inference and high-quality port selection, approaching state-of-the-art performance at a fraction of the computational cost. These results demonstrate that intelligent port selection, whether greedy or learning-based, is essential to unlocking the full multiplexing potential of multi-port \ac{FAMA} and enabling real-time deployment in \ac{RF}-chain-limited \ac{FA} receivers.

\appendices
\section{Non-Decreasing SINR Property of \ac{GFwd}}
\label{app:monotonicity}

\textit{Proposition 1:} Let $\cS \subset \{1,\ldots,P\}$ 
be a set of active ports and $p \notin \cS$. Then
\begin{equation}
\lambda_{\max}\!\big(\tilde{\bA}_{\cS \cup \{p\}},\, 
\tilde{\bB}_{\cS \cup \{p\}}\big)
\geq
\lambda_{\max}\!\big(\tilde{\bA}_{\cS},\, 
\tilde{\bB}_{\cS}\big).
\end{equation}

\begin{IEEEproof}
Let $|\cS| = t$ and $\cS' = \cS \cup \{p\}$. Define
$\bC_{\cS'} = \tilde{\bB}_{\cS'}^{-H/2}
\tilde{\bA}_{\cS'}\tilde{\bB}_{\cS'}^{-1/2}$,
so that $\lambda_{\max}(\tilde{\bA}_{\cS'},
\tilde{\bB}_{\cS'}) = \lambda_{\max}(\bC_{\cS'})$, 
and analogously $\bC_{\cS} = \tilde{\bB}_{\cS}^{-H/2}
\tilde{\bA}_{\cS}\tilde{\bB}_{\cS}^{-1/2}$.
Since $\tilde{\bA}_{\cS}$ and $\tilde{\bB}_{\cS}$ are 
principal submatrices of $\tilde{\bA}_{\cS'}$ and 
$\tilde{\bB}_{\cS'}$ sharing the same row/column indices, 
the congruence transformation $\bB^{-H/2}(\cdot)\bB^{-1/2}$ 
restricted to that index set yields $\bC_{\cS}$ as a 
principal submatrix of $\bC_{\cS'}$. As $\bA_k$ is 
positive semidefinite and $\bB_k$ is positive definite, 
both $\bC_{\cS'}$ and $\bC_{\cS}$ are Hermitian positive 
semidefinite. Applying the Cauchy interlacing 
theorem~\cite{golub2013matrix}:
\begin{equation}
\lambda_{t+1}(\bC_{\cS'}) \geq \lambda_t(\bC_{\cS}) 
= \lambda_{\max}(\bC_{\cS}),
\end{equation}
and since $\lambda_{t+1}(\bC_{\cS'}) = 
\lambda_{\max}(\bC_{\cS'})$, the result follows.
\end{IEEEproof}

\bibliographystyle{IEEEtran}
\bibliography{references/references}

\end{document}